\documentclass[letterpaper, 10 pt, conference]{ieeeconf}  

\IEEEoverridecommandlockouts                              

\usepackage{color,xcolor}
\usepackage{graphicx} 
\usepackage{booktabs}
\usepackage{multirow} 
\usepackage{threeparttable}
\usepackage{colortbl}
\usepackage{makecell}
\usepackage{amsmath} 
\usepackage{amssymb}  
\usepackage[normalem]{ulem}
\usepackage{hyperref} 

\definecolor{c2}{HTML}{FBD9BD}

\makeatletter
\renewcommand{\maketag@@@}[1]{\hbox{\m@th\normalsize\normalfont#1}}%
\makeatother

\title{\LARGE \bf
Motion-Guided Dual-Camera Tracker for Endoscope Tracking and Motion Analysis in a Mechanical Gastric Simulator
}

\author{Yuelin Zhang, Kim Yan, Chun Ping Lam, Chengyu Fang, Wenxuan Xie, Yufu Qiu, \\
Raymond Shing-Yan Tang, and Shing Shin Cheng$^{*}$
\thanks{Research reported in this work was supported in part by Innovation and Technology Commission of Hong Kong  (ITS/135/20,ITS/235/22), and The Chinese University of Hong Kong Direct Grant. The content is solely the responsibility of the authors and does not reflect the views of the sponsors.}
\thanks{Yuelin Zhang, Kim Yan, Chun Ping Lam, Wenxuan Xie, Yufu Qiu are with the Department of Mechanical and Automation Engineering, The Chinese University of Hong Kong, Hong Kong.}
\thanks{Chengyu Fang is with the Shenzhen International Graduate School, Tsinghua University, China.}
\thanks{Raymond Shing-Yan Tang is with the Department of Medicine and Therapeutics and Institute of Digestive Disease, The Chinese University of Hong Kong, Hong Kong.}%
\thanks{Shing Shin Cheng is with the Department of Mechanical and Automation Engineering, T Stone Robotics Institute, Shun Hing Institute of Advanced Engineering, Multi-Scale Medical Robotics Center, and Institute of Medical Intelligence and XR, The Chinese University of Hong Kong, Hong Kong.
        {\tt\small $^{*}$sscheng@cuhk.edu.hk}}%
}

\begin{document}

\maketitle
\thispagestyle{empty}
\pagestyle{empty}

\begin{abstract}


Flexible endoscope motion tracking and analysis in mechanical simulators have proven useful for endoscopy training. Common motion tracking methods based on electromagnetic tracker are however limited by their high cost and material susceptibility. In this work, the motion-guided dual-camera vision tracker is proposed to provide robust and accurate tracking of the endoscope tip's 3D position. The tracker addresses several unique challenges of tracking flexible endoscope tip inside a dynamic, life-sized mechanical simulator. To address the appearance variation and keep dual-camera tracking consistency, the cross-camera mutual template strategy (CMT) is proposed by introducing dynamic transient mutual templates. To alleviate large occlusion and light-induced distortion, the Mamba-based motion-guided prediction head (MMH) is presented to aggregate historical motion with visual tracking. The proposed tracker achieves superior performance against state-of-the-art vision trackers, achieving 42\% and 72\% improvements against the second-best method in average error and maximum error. Further motion analysis involving novice and expert endoscopists also shows that the tip 3D motion provided by the proposed tracker enables more reliable motion analysis and more substantial differentiation between different expertise levels, compared with other trackers. Project page: \href{https://github.com/PieceZhang/MotionDCTrack}{https://github.com/PieceZhang/MotionDCTrack}

\end{abstract}

\section{INTRODUCTION}

Gastric endoscopy is a common clinical practice that allows thorough visual inspection of the upper gastric system via a flexible endoscope. Similar to other general surgical procedures, a gastric endoscopist
must undergo proper training before performing the endoscopy on patients~\cite{kim2023simulator}. 
Mechanical simulators are typically adopted during general surgical training~\cite{hong2021simulation,king2016review,lam2024highly} with
the motion of the device (e.g. rigid laparoscopic instrument, flexible endoscope, etc.) in the simulator being analyzed
to provide quantitative and objective measurements about surgical or endoscopy skills.
However, motion analysis of flexible endoscopes in mechanical gastric simulators remains relatively primitive compared with that performed on straight rigid laparoscopic instrument~\cite{aljamal2019inexpensive}.
For example, in \cite{safavian_endoscopic_2024}, electromagnetic tracker (EMT) was attached to the endoscope tip to track its motion in an upper gastrointestinal simulator.
However, EMT is costly and its thin tethered signal cable can easily break during flexible endoscope manipulation. Besides, its tracking precision can be highly sensitive to the presence of ferromagnetic materials in the surrounding environment.
Compared with EMT, vision tracking involves low deployment cost and does not need cumbersome setup and demanding venue requirements. Therefore it has been applied 
to track rigid laparoscopic instruments manipulated in the mechanical simulator for detailed motion analysis~\cite{islam2011application,oropesa2013eva}.
However, vision tracking has thus far not been applied for motion tracking and analysis of flexible endoscope manipulation in a mechanical simulator.
Developing a robust 
visual tracking framework for flexible endoscope motion analysis using cameras installed in the inner wall of a mechanical gastric simulator thus remains an open research problem.
There are many existing works in vision tracking, including Siamese tracker \cite{li_siamrpn_2019} and its improved variants \cite{chen_siamese_2020,cheng_learning_2021,yang_siammmf_2023}.
In recent years, as transformer-based models become increasingly popular \cite{vaswani2017attention,zhang2024unified,fang2025real,fang2025integrating}, trackers based on transformers have also been proposed \cite{cui_mixformerv2_2023,gao_generalized_2023,lin_swintrack_2022,huang2024rtracker}. 
While there has been significant progress in the field of vision tracking, they are not appropriate to be directly applied for 
tracking a flexible endoscope tip inside a dynamic and realistic mechanical gastric simulator that involves many unique challenges.
For example, the manipulation of the flexible endoscope can cover a large workspace, resulting in \textbf{highly variable posture and appearance}, as well as \textbf{large occlusion}.
Furthermore, the endoscope tip features an \textbf{intense light source which can cause severe distortion} to the image captured by the cameras.

During vision-based tracking of a surgical instrument,
a multi-camera setup is usually adopted to estimate the 3D position of the target by dual-camera-based stereo matching \cite{wang_stereo_2014} or multi-camera marker-based tracking \cite{wang_surgical_2019}.
These multi-camera trackers mostly leverage rigid markers or fiducial points attached on the tracked instrument, which would require a stable and unoccluded environment to achieve the reported satisfactory performance.
For SLAM-based endoscope localization using monocular and stereo endoscope \cite{yang2022endoscope,mahmoud2017orbslam}, they are highly sensitive to sudden motion, textureless surfaces, scene variation, light distortion, etc., which can be common inside a realistic simulator.
Some SLAM methods may even require additional treatment like projected laser pattern to ensure accuracy \cite{qiu2018endoscope}.

The occlusion and light-induced distortion require additional information beyond visual features to maintain accurate target tracking.
Historical motion information has been proven to be helpful for robust tracking \cite{lin_swintrack_2022,mwikirize_timeaware_2021,yan_learningbased_2023,zhang2024mambaxctrack}
by compensating for sudden target jumps and tracking loss.
The existing works integrate motion sequence by fitting a probability model \cite{yan_learningbased_2023} or constructing plain motion token directly from original motion sequence \cite{lin_swintrack_2022} but fail to explore the long-range interrelationship within the time domain.
Structured state space sequence models (SSMs) \cite{kalman1960new} draw considerable attention due to their long-range modeling ability, especially the Mamba \cite{gu_mamba_,gu2021combining}, which introduces the selective scan mechanism to model long-range relationships in an input-dependent manner.
Beyond its original version, variations make a series of advancements, including special scanning strategies \cite{pei2024efficientvmamba,tang2024scalable}, transformer combinations \cite{lieber2024jamba}, bidirectional structures \cite{zhu_vision_2024,zhang2024motion}, etc.

In this work, a motion-guided dual-camera tracker is presented to enable for the first time tracking of a flexible endoscope tip in a life-sized mechanical gastric simulator and thus its 3D motion analysis.
Our proposed tracking framework is designed to address the challenges of large appearance variation of the endoscope tip, its temporary occlusion and disappearance, and significant distortion by the light source from the endoscope. 
First, instead of using template updating or template-free strategy to adapt to appearance variation \cite{fu2021stmtrack,sun_fast_2020}, a \textbf{cross-camera mutual template strategy (CMT)} is proposed as a dual-camera integration scheme to make full use of mutual information from coupled cameras.
CMT enables the tracking system to benefit from the mutual template from synchronized frames in the coupled dual cameras.
As a result, tracking a target with a volatile appearance can be simplified into feature matching between dynamic transient mutual templates from dual cameras, leading to better performance than marker-based and SLAM methods.
CMT can also improve 3D tracking accuracy by introducing dual-camera tracking consistency.
Second, beyond the existing methods without modeling motion interrelationship, a \textbf{Mamba-based motion-guided prediction head (MMH)} is incorporated to construct bidirectional motion tokens from long-range temporal dependencies, enabling robust tracking during target disappearance and under significant image distortion. 
Our proposed tracker achieves robust and accurate dual-camera tracking of the flexible endoscope tip with highly variable postures under a noisy environment in the mechanical simulator, outperforming state-of-the-art trackers.
The significance of more accurate tracking is also reflected in the more accurate motion analysis, allowing differentiation between experts and novices.
The contributions of our work are threefold:
\begin{itemize}
    \item The dual-camera-based cross-camera mutual template strategy (CMT) is proposed to adapt to the variable appearance while enhancing dual-camera tracking consistency. It is the first time that a dual-camera integration strategy has been proposed for a multi-camera tracker.
    \item The Mamba-based motion-guided prediction head (MMH) is proposed to integrate historical motion, achieving robust tracking against target disappearance and strong distortion. This is also, to our knowledge, the first time Mamba has been adopted for motion modeling and object tracking in a medical scenario.
     \item Extensive experiments show that the proposed tracker outperforms existing trackers in both 2D and 3D evaluation, as well as motion analysis. Further ablation studies also demonstrate the effectiveness of the proposed MMH and CMT.
\end{itemize}

\begin{figure*}
\centering
\includegraphics[width=\textwidth]{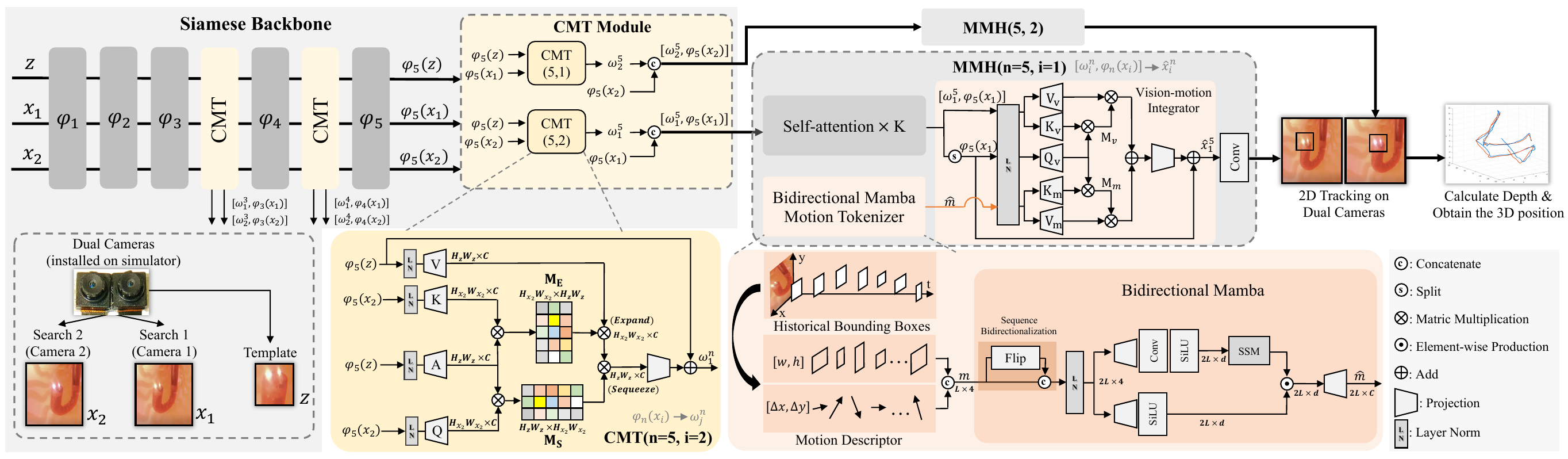}
\caption{
Structure overview. 
$\varphi_1$ to $\varphi_5$ denote the layers in the Siamese ResNet \cite{he2016deep} backbone, $\varphi_n(x_i)$ and $\varphi_n(x_j)$ are the intermediate output from backbone, where $n\in \{3,4,5\}$, $\{i,j\} = \{1,2\}$. 
Three CMTs are cascaded behind $\varphi_3$, $\varphi_4$, and $\varphi_5$. Each of the three CMTs is then followed by an MMH.
For simplicity, the figure only shows details in CMT(5,2) and MMH(5,1). All CMTs and MMHs follow the same workflow.
}
\label{fig_overview}
\end{figure*}

\section{Methodology}

\subsection{Overview}
As shown in Fig.~\ref{fig_overview}, the Siamese ResNet \cite{he2016deep} backbone receives two search maps $x_1$ and $x_2$ from dual cameras and a template map $z$ (from camera 1 by default). It then extracts features following a similar workflow with existing Siamese trackers \cite{lin_swintrack_2022,chen2022siamban}.
The CMT and MMH are placed after backbone stages $\varphi_n$ ($n\in \{3,4,5\}$).
CMT aggregates features $\varphi_n(x_i)$ from the coupled camera into the original template $\varphi_n(z)$, generating mutual templates $\omega^n_j$ for $\varphi_n(x_j)$ ($\{i,j\} = \{1,2\}$).
In the following MMH, the concatenated $[\omega^n_i, \varphi_n(x_i)]$ goes through multi-head self-attention \cite{vaswani_attention_2017}, and then goes into vision-motion integrator to integrate historical motion from Mamba bidirectional motion tokenizer.
The 2D tracking result is obtained by averaging prediction maps from three MMHs at $n=3,4,5$ stages.
The depth of the tracked target is then estimated based on stereo disparity (difference of target position in two cameras) \cite{hamzah2016literature}, which is given by
\begin{equation} d=\frac{B\cdot f}{D\cdot d_{x} }, \label{eq_d} \end{equation}
where $B$ is the baseline distance between the center of binocular cameras, $f$ is the camera's focal length, $d_{x}$ is the physical pixel size on the camera sensor along the x direction, and $D$ is the disparity. 
The 3D position of the endoscope tip can then be obtained.

\subsection{Cross-camera Mutual Template Strategy (CMT)}
Since the flexible endoscope tip has highly variable posture and appearance, the original template from the initial frame may not be always informative throughout the whole procedure.
Furthermore, the tracking consistency between dual cameras is important for the accuracy of stereo disparity and 3D position \cite{hamzah2016literature}.
By assuming the transient features from dual cameras have high consistency, CMT dynamically generates mutual templates for each camera by aggregating the synchronized frames from its coupled camera, as shown in Fig.~\ref{fig_overview}.
CMT relies on the proposed \textbf{anchored expansion-squeeze cross-attention}, given by:

\begin{align} 
    & M_E = Softmax(\textbf{K} \cdot \textbf{A}^T / \sqrt{C}), \\
    & M_S = Softmax(\textbf{A} \cdot \textbf{Q}^T / \sqrt{C}), \\
    & \omega^n_j = Linear(M_S \cdot (M_E \cdot \textbf{V}))+\varphi_n(z), 
\end{align}


\noindent where $C$ denotes the embedded dimension ($C=256$ in this paper).
$\textbf{A}=Proj(LN(\varphi_n(z)))$ is an additional anchor projection besides the standard $\textbf{Q}=Proj(LN(\varphi_n(x_i))),\textbf{K}=Proj(LN(\varphi_n(x_i))),\textbf{V}=Proj(LN(\varphi_n(z)))$ projections, where $LN(\cdot)$ refers to layer normalization.
Using $\textbf{A}$ as an intermediate transformation between 
$\varphi_n(z)$ and $\varphi_n(x_i)$ in different size, the expansion attention map $M_E\in \mathbb{R}^{H_{x_i}W_{x_i} \times H_zW_z}$ and squeeze attention map $M_S\in \mathbb{R}^{H_zW_z \times H_{x_i}W_{x_i}}$ are obtained.
The anchored expansion-squeeze operation is performed by multiplying $\textbf{V}$ with $M_E$ and $M_S$ successively, where $\textbf{V}$ is expanded into a larger embedded space with richer representation and then squeezed back to its original size.
Different from the existing expansion-squeeze that models self-attention within a single size and expands in channel dimension \cite{li2021medical}, the proposed workflow models positional cross-attention between template $z$ and search $x$ in different sizes while keeping the output size unchanged with $z$.
It enables modeling positional attention between maps with different sizes while enlarging the intermediate projection space.
The obtained mutual template $\omega^n_j$ is then concatenated with $\varphi_n(x_j)$ for tracking prediction.

CMT enables the potential of generalizing to unseen features since the mutual templates are dynamically obtained from coupled cameras in a training-data-agnostic way.
The transient mutual templates guarantee timeliness and informativeness, addressing the appearance variation problem and ensuring dual-camera tracking consistency.

\subsection{Mamba-based Motion-guided Prediction Head (MMH)}
The MMH receives search map $\varphi_n(x_i)$ and its mutual template $\omega^n_i$.
As shown in Fig.~\ref{fig_overview},  the concatenated input map $[\omega^n_i,\varphi_n(x_i)]$ is first processed by $K$ cascaded multi-head self-attention modules \cite{vaswani_attention_2017} ($K=6$).
To construct the motion token, instead of directly using bounding box position \cite{lin_swintrack_2022}, in this paper, the historical bounding boxes are first converted from absolute position in the image coordinate to relative descriptors in the bounding box coordinate.
This conversion enhances generalizability by replacing absolute value with relative local parameters.
The obtained low-level local descriptors contain box width and height ($w, h$) and displacement in $x$ and $y$ direction ($\Delta x,\Delta y$) along the time domain.
They are then concatenated to learn the latent internal relationship within the descriptor.

The following \textbf{bidirectional Mamba block} models the long-range dependencies along time.
Since the original Mamba is unidirectional, the bidirectionalization is first performed as shown in Fig.~\ref{fig_overview} to expand the raw sequence $m\in\mathbb{R}^{L\times 4}$ into bidirectional form with a size of $2L\times 4$.
After a layer normalization, embedded maps with a size of $2L\times d$ are obtained, where $d=128$ and $L=240$ in this paper. Here SiLU activation \cite{elfwing2018sigmoid} is used.
SSM is defined by linear Ordinary Differential Equations (ODEs) given by 
\begin{align}
    & h^\prime(t)=\textbf{A}h(t)+\textbf{B}x(t), \\
    & y(t)=\textbf{C}h(t),
\end{align}
where $\textbf{A}\in \mathbb{R}^{N\times N}$ is the state matrix, $\textbf{B}\in \mathbb{R}^{N\times 1}$ and $\textbf{C}\in \mathbb{R}^{1\times N}$ are projection matrices. 
It maps the input sequence $x(t)\in\mathbb{R}^N$ to output $y(t)\in\mathbb{R}^N$ with latent states $h(t)\in\mathbb{R}^N$. 
These linear ODEs are then discretized as 
\begin{align}
& h_t=\bar{\textbf{A}}h_{t-1}+\bar{\textbf{B}}x_t, \\
& y_t=\textbf{C}h_t.     
\end{align}
The discritized matrices $\bar{\textbf{A}}$ and $\bar{\textbf{B}}$ are given by $\bar{\textbf{A}}=exp(\mathbf{\Delta}\cdot \textbf{A})$ and $\bar{\textbf{B}}=(\mathbf{\Delta}\cdot \textbf{A})^{-1}(exp(\mathbf{\Delta}\cdot\textbf{A})-I)\cdot(\mathbf{\Delta} \textbf{B})$, where $\mathbf{\Delta}$ is the discretization step size.
Here selective scan SSM \cite{gu_mamba_} is adopted.
It improves the traditional SSMs by parameterizing the SSM based on input, where the parameters $\mathbf{\Delta},\bar{\textbf{B}},\textbf{C}$ are obtained from projections of the input sequence.
Finally, the embedded maps are projected into bidirectional motion token $\hat{m}\in\mathbb{R}^{2L\times C}$.
Different from the bidirectional structure in \cite{zhu_vision_2024} with two independent SSMs, the proposed structure models the whole bidirectional sequence in a single SSM to better adapt to the motion hints.

The motion token $\hat{m}$ is then integrated with the vision feature using the proposed \textbf{vision-motion integrator}, which is a multi-${KV}$ cross-attention as shown in Fig.~\ref{fig_overview}.
This operation is given by:

\begin{align} 
    & M_v = Softmax(\mathbf{Q_v} \cdot \mathbf{K_v}^T / \sqrt{C}), \\
    & M_m = Softmax(\mathbf{Q_v} \cdot \mathbf{K_m}^T / \sqrt{C}), \\
    & \hat{x}^n_i = Linear(M_v \cdot \mathbf{V_v} + M_m \cdot \mathbf{V_m}) + \varphi_n(x_i), 
\end{align}



\noindent where $M_v$ and $M_m$ are attention maps for visual feature and motion hints, respectively. 
$\mathbf{Q_v}=Proj(LN(\varphi_n(x_i)))$, $\mathbf{K_v}=Proj(LN([\omega^n_i,\varphi_n(x_i)]))$, and $\mathbf{V_v}=Proj(LN([\omega^n_i,\varphi_n(x_i)]))$ are the projection with visual information. $\mathbf{K_m}=Proj(LN(\hat{m}))$ and $\mathbf{V_m}=Proj(LN(\hat{m}))$ are the projection containing motion prompts.
With the integrator, the vision feature map is losslessly aggregated with the historical motion hints, without losing the self-contained positional embedding. 

With MMH, the historical motion is tokenized as non-visual hints for robust tracking.
It is helpful when the target is temporarily lost due to occlusion or light disturbance.

\section{Experiments and Results}

\begin{figure}
\centering
\includegraphics[width=1.0\linewidth]{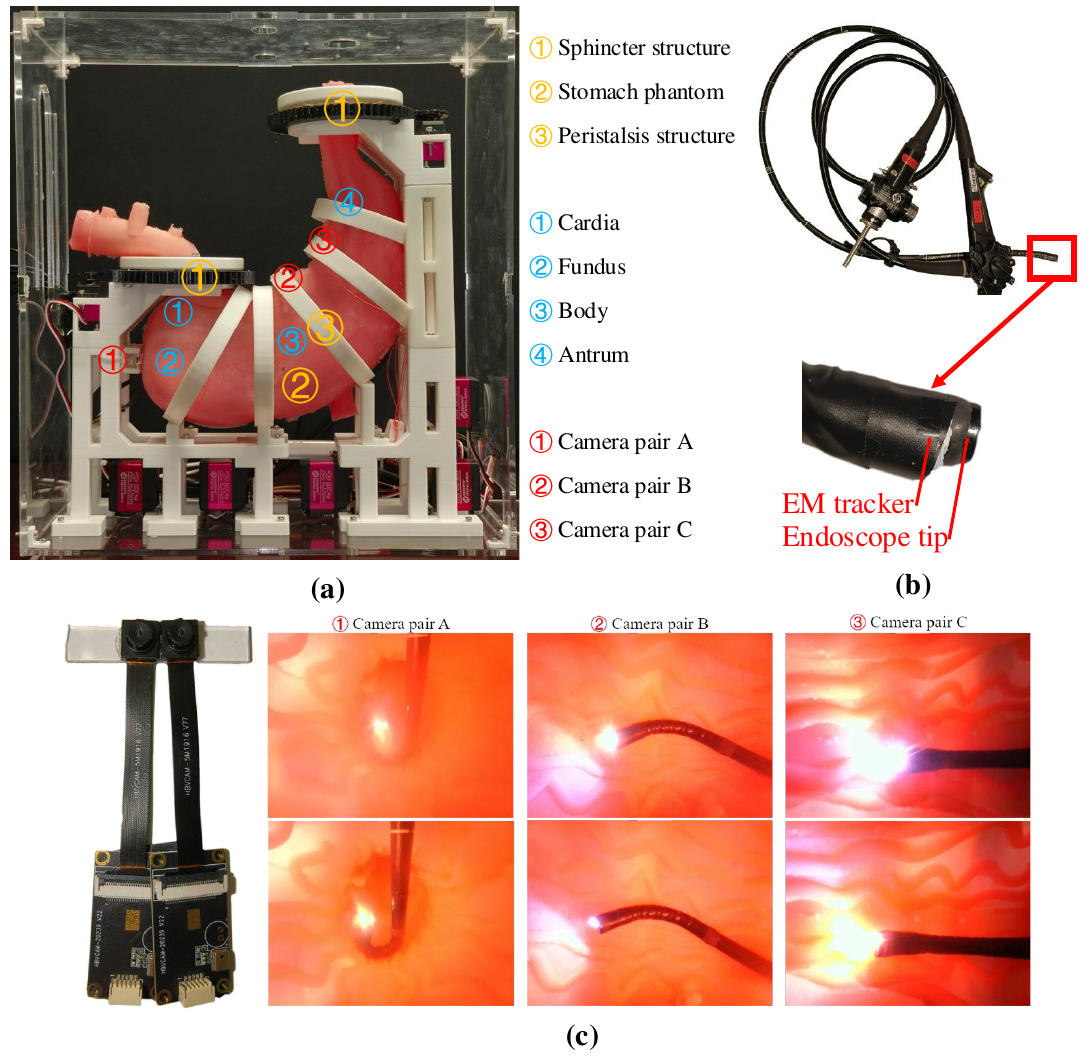}
\caption{
Experiment setup.
(a) Self-developed mechanical gastric simulator and installation of dual-camera tracking devices. 
(b) Flexible gastric endoscope with EMT affixed at its tip to provide the 3D ground truth. 
(c) Dual camera pairs used in this work and image examples collected by different camera pairs.
}
\label{fig_simulator}
\vspace{-8mm}
\end{figure}

\subsection{Experiment Setup and Dataset Collection}
As shown in Fig.~\ref{fig_simulator}(a), a customized mechanical gastric simulator was developed as a realistic simulating platform for endoscopy training. The salmon-color, highly distensible silicone stomach phantom has a thin wall and two openings with sphincters.
It has four main parts, namely cardia, fundus, body, and antrum. 
The inner side of the model is lined with rugae to imitate the actual stomach wall.
It has a sphincter movement structure and a peristalsis actuation structure to simulate the dynamic behavior of the human stomach.
The flexible endoscope enters the phantom through the esophageal sphincter before the phantom is inflated and the integrated peristalsis mechanism is activated to simulate peristalsis motion along the stomach wall.
Three pairs of calibrated binocular cameras (Fig.~\ref{fig_simulator}(c)) were installed on the inside of the phantom wall as shown in Fig.~\ref{fig_simulator}(a), where pair A was at the fundus, and pair B and C were on the lesser curvature of the stomach body. 
Note that such dual-camera pairs can be constructed at a low cost using cheap cameras (HBV-5M2118 by Huiber Vision Technology, $\sim$15 USD).

An Olympus GIF-FQ260Z endoscope was manipulated inside the simulator during data collection.
48 videos (1280$\times$720, 30 FPS) of endoscope manipulation were finally acquired.
Each video contains light disturbance by tip light, occlusion by scope retroflexion, and appearance variation by large maneuvering.
By downsampling these videos to 2 FPS, a dataset containing 14530 frames is obtained.
The 2D ground truth (bounding boxes of the endoscope tip) was annotated by only one annotator to eliminate disagreement.
The 3D position ground truth was measured by an EMT affixed on the endoscope tip (Fig.~\ref{fig_simulator}(b)).
During manipulation, the simulator was kept away from disturbance. The EM tracker has been verified to have an RMSE of 0.76 mm.

To prevent overfitting and evaluate the generalizability, a 6-fold cross-validation is performed based on a random video-level data splitting strategy.
The dataset is split into 6 subsets on a video basis, where each subset contains 8 videos. 
The final result is reported by repeating the iteration six times and averaging the result from each iteration.
The program was implemented with PyTorch.
The proposed tracker and all comparison methods were trained on the dataset by four NVIDIA RTX 4090 GPUs. 
All the baselines adopt the same training scheme (150 epochs, batch size of 24).
An AdamW optimizer is applied with a weight decay 1e-4, a learning rate 5e-4, and a backbone learning rate 5e-5. The learning rate is dropped by 10 after 100 epochs.
During training, a short historical motion segment with the current image is taken to train the MMH.
Common augmentations, including position shifting, scaling, blur, flip, and color jitter, are applied.
Gaussian noise is added to the motion segment.
Since no similar \textbf{labeled dual-camera tracking dataset} is available, tests are only performed on our dataset.

\begin{table}
\begin{center}
\caption{
Endoscope tracking comparison. Both results of 2D and 3D metrics are reported based on the averaged values from 6-fold cross-validation. 
The methods with the best and second-best performance are noted in {\color[HTML]{FF0000}red} and {\color[HTML]{00B0F0}cyan}.
}\label{tab_result}
\setlength{\tabcolsep}{1.5mm}{
\scalebox{0.96}{
\begin{tabular}{l|c|c|c|c|c}
\toprule
\multirow{2}{*}{Method} & \multicolumn{2}{c|}{2D Metrics (\%)}  & \multicolumn{3}{c}{3D Metrics (mm)} \\
 & \cellcolor{c2!50}SUC $\uparrow$ & \cellcolor{c2!50}PRE $\uparrow$ & \cellcolor{c2!50}Avg err. $\downarrow$ & \cellcolor{c2!50}Max err. $\downarrow$ & \cellcolor{c2!50}SD $\downarrow$ \\
\midrule
SiamRPN++ \cite{li2019siamrpn++}& 62.5& 61.7& 14.70& 423.26& 17.58\\
SiamBAN \cite{chen2022siamban} & 66.0& 68.2& 8.72& 310.84& 13.94\\
SiamAttn \cite{yu2020deformable} & 65.1& 62.8& 9.41& 92.96& 8.47\\
STMTrack \cite{fu2021stmtrack} & 68.3& 69.6& \color[HTML]{00B0F0}8.59& 90.70& 8.71\\
SwinTrack \cite{lin_swintrack_2022} & 73.9& 74.2& 10.09& 63.51&\color[HTML]{00B0F0}7.61\\
MixFormerV2 \cite{cui_mixformerv2_2023} & \color[HTML]{00B0F0}76.0& \color[HTML]{00B0F0}76.5& 8.83&\color[HTML]{00B0F0}57.04& 10.82\\
\textbf{Ours} &\color[HTML]{FF0000}78.9& \color[HTML]{FF0000}79.1& \color[HTML]{FF0000}5.13& \color[HTML]{FF0000}16.01& \color[HTML]{FF0000}3.84\\
\bottomrule
\end{tabular}
}
}
\end{center}
\vspace{-4mm}
\end{table}

\begin{figure}
\centering
\includegraphics[width=1.0\linewidth]{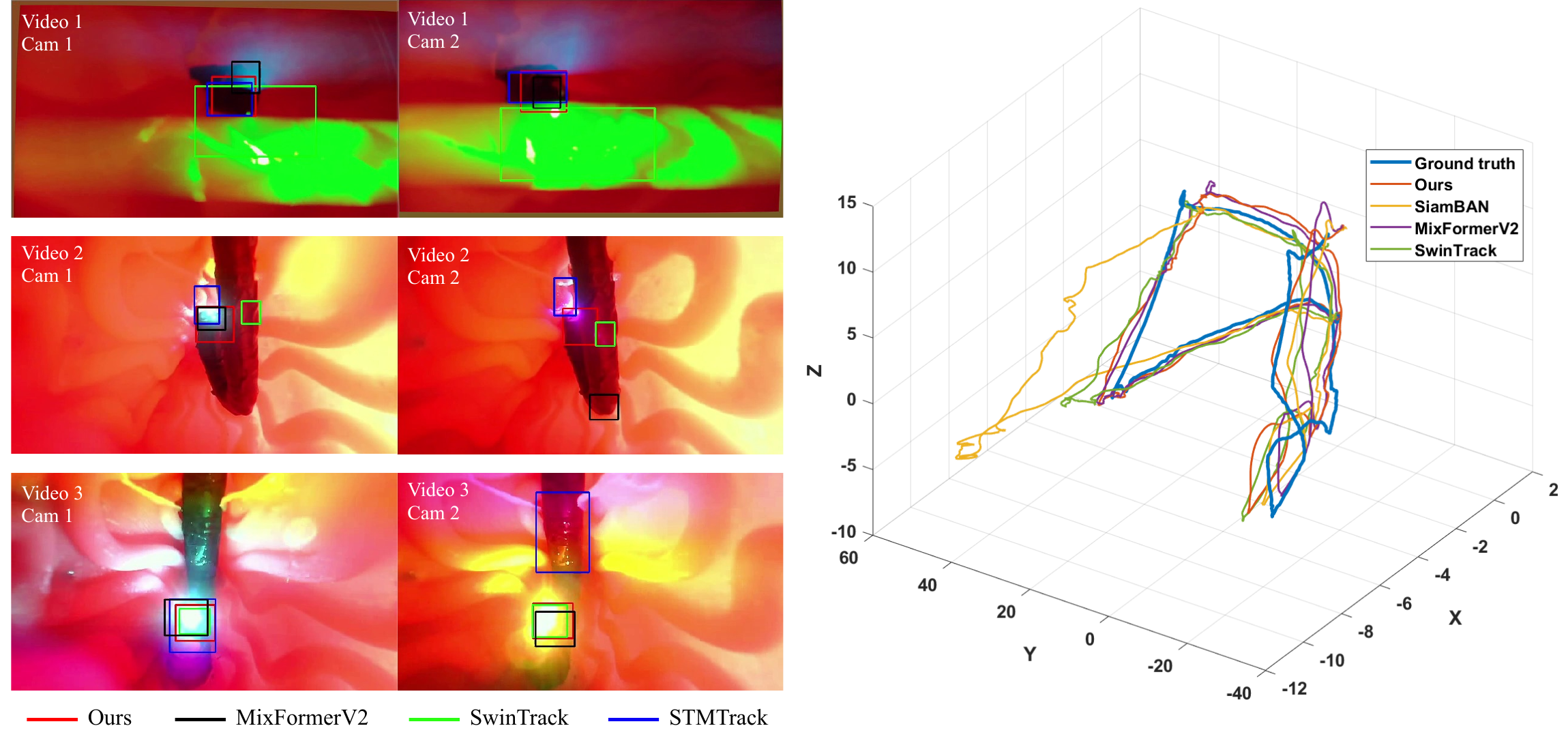}
\caption{
Left: Demonstration of the dual-camera tracking comparison. Our tracker not only achieves the most accurate tracking under multiple disturbances but also has the best tracking consistency across the dual cameras (See supplementary video for more tracking demonstration).
Right: 3D motion trajectory ground truth measured by EMT and estimated 3D motion from different methods. 
}
\label{fig_result}
\end{figure}

\subsection{Results}
The proposed tracker is compared against several state-of-the-art trackers, as shown in Tab.~\ref{tab_result}, including the latest transformer-based tracker \cite{cui_mixformerv2_2023}, tracker with motion token \cite{lin_swintrack_2022}, tracker with template-free strategy STMtrack \cite{fu2021stmtrack}, classical Siamese tracker \cite{li2019siamrpn++}, etc.
All methods first perform 2D tracking on each image of dual cameras and estimate 3D position using stereo disparity.
The results show that the proposed tracker achieves SOTA performance with a 78.9 success rate (SUC) and 79.1 precision (PRE), outperforming the second-best method MixFormerV2 by 2.9 and 2.6.
It shows that the proposed framework is especially effective in dual-camera endoscope tracking scenario, successfully addressing the challenges inside the simulator, including appearance variation, large occlusion, and light-induced image distortion, as shown in the tracking demonstration\footnote{More tracking demonstration is provided in the supplementary video.} given in Fig.~\ref{fig_result}.
Both quantitative and qualitative results show that the proposed tracker leads to better performance with more accurate tracking than the comparison methods.
Furthermore, the proposed tracker has better tracking consistency between dual cameras, i.e., tracked bounding boxes in both two cameras strictly refer to the same target, as shown in Fig.~\ref{fig_result}. 
This feature improves the accuracy of the stereo disparity, which can be helpful for 3D position estimation since the depth calculation is directly related to the stereo disparity.
This advantage can be observed from 3D metrics in Tab.~\ref{tab_result}, where three metrics are used, including average error, maximum error, and standard deviation (SD).
Our tracker outperforms the other methods significantly in all three metrics, achieving 42\%, 72\%, and 65\% improvement respectively against the second-best method.
The estimated 3D trajectories shown in Fig.~\ref{fig_result} also indicate that the 3D trajectory obtained from our tracker has less noise and deformation.
It is worth noting that the proposed tracker keeps efficiency while achieving SOTA, which runs dual-camera tracking on an NVIDIA RTX 4090 GPU at a real-time speed of 34.2 FPS.

\begin{table}
\begin{center}
\caption{
Ablation study on MMH and CMT. Results are reported by 6-fold cross-validation.
}\label{tab_ablation}
\setlength{\tabcolsep}{1.0mm}{
\scalebox{0.82}{
\begin{threeparttable}
    \begin{tabular}{l|l|l|l|l|l}
    \toprule
    \multirow{2}{*}{Method} & \multicolumn{2}{c|}{2D Metrics (\%)}  & \multicolumn{3}{c}{3D Metrics (mm)} \\
     & \cellcolor{c2!50}SUC $\uparrow$ & \cellcolor{c2!50}PRE $\uparrow$ & \cellcolor{c2!50}Avg err. $\downarrow$ & \cellcolor{c2!50}Max err. $\downarrow$ & \cellcolor{c2!50}SD $\downarrow$ \\
    \midrule
    Baseline & 70.8& 70.5 & 10.52& 206.48& 11.20\\
    \midrule
    Baseline + MMH$^{*}$ & \textbf{74.7 {\tiny (+3.9)}}& 75.2 {\tiny (+4.7)}& \textbf{9.30 {\tiny (-1.22)}}& 66.21 {\tiny (-140.27)}& \textbf{5.70 {\tiny (-5.5)}}\\
    Baseline + MMH ($L=30$) & 71.0 {\tiny (+0.2)}& 70.1 {\tiny (-0.4)}& 11.59 {\tiny (+1.07)}& \textbf{46.40 {\tiny (-160.08)}}& 9.96 {\tiny (-1.24)}\\
    Baseline + MMH ($L=600$) & 74.6 {\tiny (+3.8)}& \textbf{75.4 {\tiny (+4.9)}}& 9.69 {\tiny (-0.83)}& 90.52 {\tiny (-115.96)}& 7.35 {\tiny (-3.85)}\\
    \midrule
    Baseline + CMT & 76.1 {\tiny (+5.3)}& 76.6 {\tiny (+6.1)}& 7.04 {\tiny (-3.48)}& 22.59 {\tiny (-183.89)}& 9.16 {\tiny (-2.04)}\\
    SwinTrack \cite{lin_swintrack_2022} + CMT & 75.0 {\tiny (+1.1)}& 74.4 {\tiny (+0.2)}& 8.62 {\tiny (-1.47)}& 40.92 {\tiny (-22.59)}& 7.57 {\tiny (-0.04)}\\ 
    MixFormerV2 \cite{cui_mixformerv2_2023} + CMT & 77.5 {\tiny (+1.5)}& 78.6 {\tiny (+2.1)}& 6.02 {\tiny (-2.81)}& 25.10 {\tiny (-31.94)}& 11.51 {\tiny (+0.69)}\\ 
    \midrule
    Baseline + MMH + CMT & 78.9 {\tiny (+8.1)}& 79.1 {\tiny (+8.6)}& 5.13 {\tiny (-5.39)}& 16.01 {\tiny (-190.47)}& 3.84 {\tiny (-7.36)}\\
    \bottomrule
    \end{tabular}

    \begin{tablenotes}
        \footnotesize
        \item $^{*}$ default value of $L$ in this paper is 240
    \end{tablenotes}
    
\end{threeparttable}
}
}
\end{center}
\end{table}

\subsection{Ablation Study}
During the ablation study, the baseline without MMH refers to the model with the Mamba motion tokenizer removed, where the MMH is then degraded into a commonly used simple prediction head with self-attention and cross-attention.
As shown in Tab.~\ref{tab_ablation}, the model with MMH or CMT has significant improvement over the baseline in both 2D and 3D metrics.
The baseline with MMH improves the SUC and PRE by 3.9 and 4.7. And it significantly reduces the SD by 49\%. 
This improvement demonstrates that the historical motion hints can work as an effective prompt for tracking under a challenging environment with occlusion and disturbance. 
MMH also helps the tracker avoid large errors that may be caused by target disappearance, which can be observed from the improvement in maximum error.
Ablation on the hyper-parameter $L$ is also conducted.
The results show performance degradation on the model with both longer sequences ($L=600$) and shorter sequences ($L=30$), compared with the default configuration $L=240$.

The evaluation of the baseline with CMT also reports improvements among all involved metrics, showing that with the integration of CMT, the tracker is enhanced by adapting to appearance variation with dynamic mutual templates.
Significant improvements are observed in two 3D metrics, which are 33\% for average 3D error and 89\% for maximum 3D error.
It demonstrates that the proposed CMT can bring enhancement to 3D position estimation based on dual-camera tracking, leveraging the cross-camera mutual templates to ensure cross-camera tracking consistency.
Additional generalization tests were performed by applying the CMT strategy on two SOTA trackers, namely SwinTrack~\cite{lin_swintrack_2022} and MixFormerV2 \cite{cui_mixformerv2_2023}.
The following improvement shows the generalizability of the proposed CMT.

\subsection{Discussion}
Compared with STMTrack \cite{fu2021stmtrack}, which adopts a template-free strategy by using embedded features of historical frames as templates, the mutual template strategy applied in CMT uses transient features from coupled cameras as templates to guarantee timeliness and informativeness while avoiding error accumulation, thus outperforming STMTrack.
SwinTrack \cite{lin_swintrack_2022} also adopts motion token, but it directly uses plain motion sequence as motion token without modeling time-domain interdependencies, resulting in less benefit from motion information.
And it fails to extract lower-level motion descriptors to eliminate absolute information, which may pose potential harm to its generalizability.

The proposed tracker achieves SOTA in both 2D and 3D evaluations.
According to Tab.~\ref{tab_result}, the improvement in 3D metrics is much more significant than in 2D metrics.
The reason is that the proposed CMT can not only tackle appearance variation by dynamic transient mutual templates but also introduce binocular visual constraints into dual-camera tracking, thus ensuring tracking consistency between two cameras.
This improvement greatly improves the accuracy of 3D position estimation that relies on stereo disparity.

\begin{table}
\begin{center}
\caption{Motion analysis according to motion metrics in \cite{oropesa2013eva}. The expertise level differentiation is reported by statistical significance analysis.
Detail descriptions of motion metrics are provided in Tab.~\ref{tab_motionmetrics}.
}
\label{tab_eval}
\scalebox{0.78}{
\begin{threeparttable}
    \begin{tabular}{l|l|l|l|l|l|l|l|l}
    \toprule
    & & \cellcolor{c2!50}\makebox[0.03\textwidth][l]{T} & \cellcolor{c2!50}\makebox[0.03\textwidth][l]{IT} & \cellcolor{c2!50}\makebox[0.03\textwidth][l]{PL} & \cellcolor{c2!50}\makebox[0.03\textwidth][l]{S} & \cellcolor{c2!50}\makebox[0.03\textwidth][l]{A} & \cellcolor{c2!50}\makebox[0.03\textwidth][l]{MS} & \cellcolor{c2!50}\makebox[0.03\textwidth][l]{EOV}\\
    \midrule
    \multirow{3}{*}{\makecell[l]{Ours}}   & expert & 125  & 18.41  & 8.09  & 31.17  & 12.49  & 13.41  & 0.0056 \\
                            & novice & 217 &  10.28  & 27.74  & 46.32  & 15.90  & 15.12  & 0.0036 \\
                            & \textit{p} value$^{*}$ & \textbf{0.003}& \textbf{0.011}  & \textbf{0.003}& \textbf{0.003}& \textbf{0.005} & 0.25 & \textbf{0.029} \\
    \midrule
    \multirow{3}{*}{\makecell[l]{MixFormerV2 \\ \cite{cui_mixformerv2_2023}}}   
    & expert & 125  & 16.05  & 14.51& 41.87& 15.02& 11.59& 0.0062\\
    & novice & 217 &  15.52& 31.06& 50.82& 16.94& 10.04& 0.0039\\
    & \textit{p} value$^{*}$ & \textbf{0.003}& 0.19& \textbf{0.008}& \textbf{0.012}& 0.059& 0.30& \textbf{0.024}\\
    \midrule
    \multirow{3}{*}{\makecell[l]{SwinTrack \\ \cite{lin_swintrack_2022}}}   
    & expert & 125  & 21.06& 11.71& 38.63& 13.76& 12.80& 0.0070\\
    & novice & 217 &  13.41& 29.50& 49.02& 16.11& 13.29& 0.0056\\
    & \textit{p} value$^{*}$ & \textbf{0.003}& \textbf{0.015}& \textbf{0.003}& \textbf{0.005}& \textbf{0.007}& 0.45& 0.060\\
    \bottomrule
    \end{tabular}
    
    \begin{tablenotes}
        \footnotesize
        \item $^{*}$ The statistical significance (\textit{p} value) is given by a Mann-Whitney \textit{U}-test, where significant differences at the \textit{p} $\leq$ 0.05 level are indicated in bold.
    \end{tablenotes}
    
\end{threeparttable}
}
\end{center}
\end{table}

\begin{figure}
\centering
\includegraphics[width=0.95\linewidth]{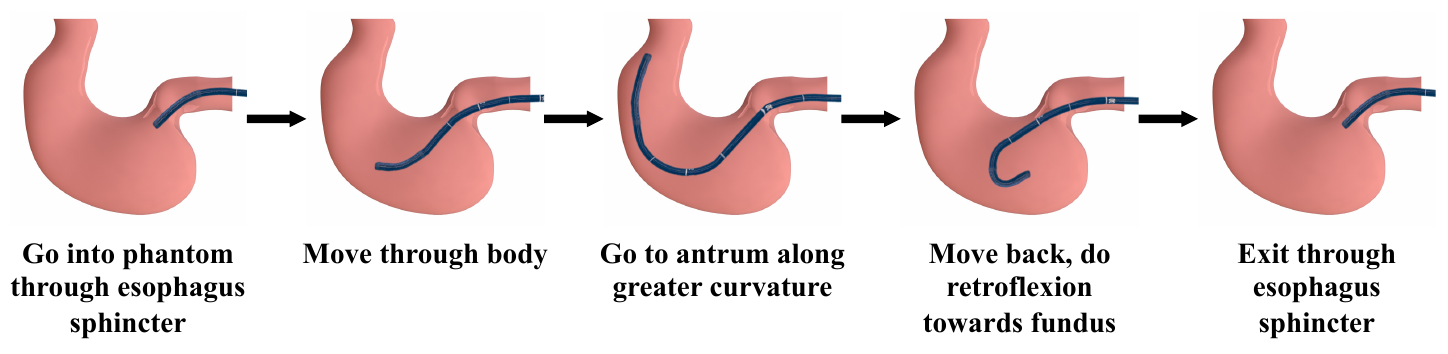}
\caption{
Demonstration of the procedures done during motion analysis.
}
\label{fig_motion}
\end{figure}

\subsection{Motion Analysis}
In addition to the dataset that has been used for training and testing, data for motion analysis tests were collected by inviting expert and novice surgeons (expertise defined according to the number of procedures performed based on the ASGE standards) to perform a series of clinically-driven endoscopy procedures in the gastric simulator, including navigating along anatomical landmarks and perform difficult maneuvers such as retroflexion (Fig.~\ref{fig_motion}). 
A total of 15 trials by three experts (more than 1000 endoscopy cases performed) and 36 trials by six novices (less than 130 endoscopy cases performed) are included. 
The 3D motion trajectories of the endoscope tip were acquired by the proposed tracker and the other two SOTA trackers.
The motion analysis is then given by motion metrics~\cite{oropesa2013eva} (Tab.~\ref{tab_motionmetrics}), including T (time), IT (idle time percentage), PL (average path length), S (average speed), A (average acceleration), MS (motion smoothness), and EOV (economy of volume). 

The results are shown in Tab.~\ref{tab_eval}.
To assess the performance of different trackers on motion analysis, i.e., whether the results from different trackers can accurately reflect levels of proficiency and distinguish between participants with different expertise levels, expertise level differentiation between experts and novices is evaluated by conducting statistical significance tests.
Ideally, the tracker with better performance can lead to a more significant differentiation, since it tends to have less trajectory distortion and jerk.
A Mann-Whitney \textit{U}-test is performed and the statistical significance is given by the \textit{p} value, as shown in Tab.~\ref{tab_eval}.
For our tracker, statistical significance (\textit{p} $\leq$ 0.05) was found in 6 out of 7 metrics.
As for the tracker with the second-best accuracy MixFormerV2 \cite{cui_mixformerv2_2023}, the statistical significance is only reported in 4 out of 7 metrics. 
For SwinTrack \cite{lin_swintrack_2022}, the number is 5 out of 7.
This comparison shows that tracking accuracy and robustness have a notable impact on motion analysis and skill differentiation outcomes. 
Trackers with worse tracking performance also tend to have worse motion analysis due to their inaccurate and indistinguishable tracking trajectories.
By integrating MMH and CMT, our tracker successfully address the unique challenges of flexible endoscope tracking, resulting in superior tracking performance that leads to more accurate motion analysis and skill differentiation.

\section{CONCLUSIONS}
In this paper, a motion-guided dual-camera tracker with CMT and MMH is proposed for vision-based tracking of the endoscope tip inside a mechanical gastric simulator to allow endoscopy motion analysis.
The tracker achieves SOTA performance, enabling reliable and accurate tip 3D position feedback.
Since no other dual-camera target tracking dataset is publicly available, the current evaluation only involves our self-collected dataset. In future work, more experiments will be performed on open-world datasets for more comprehensive validation on generalization.
Endoscopic skill training and evaluation involving a large cross-institution cohort will also be conducted based on the proposed tracker.
The proposed CMT and MMH modules may also be applied to robustly track flexible surgical instruments and integrated into a closed-loop control framework for flexible robotic surgery under a volatile environment.




\section*{APPENDIX}
The metrics of motion analysis are shown in Tab.~\ref{tab_motionmetrics}.

\begin{table}[htbp]
  \centering
  \caption{Description of motion metrics \cite{oropesa2013eva}.}
  \label{tab_motionmetrics}
    \resizebox{\linewidth}{!}{
    \begin{tabular}{llp{9.0em}c}
    \toprule
    \cellcolor{c2!50} Metrics & \cellcolor{c2!50} Units & \multicolumn{1}{l}{\cellcolor{c2!50} Definition} & \cellcolor{c2!50} Formulae \\
    \midrule
    \makecell[l]{T \\ (time)} & s     & Total time to perform a task & T \\
    \midrule
    \makecell[l]{IT \\ (idle time percentage)} & \%    & Percentage of time where the instrument is considered to be still & $ {\scriptscriptstyle \frac{|\Im|}{T}: \Im=\left\{t \in(0, \ldots T) \mid \sqrt{\left(\frac{d x(t)}{d t}\right)^2+\left(\frac{d y(t)}{d t}\right)^2+\left(\frac{d z(t)}{d t}\right)^2} \leq 5\right\}} $ \\
    \midrule
    \makecell[l]{PL \\ (average path length)} & m     & Total path covered by the instrument & $\int_{t=0}^T \frac{d|r(t)|}{d t} d t$ \\
    \midrule
    \makecell[l]{S \\ (average speed)} & mm/s  & Rate of change of the instrument’s position & $ \frac{1}{T} \int_{t=0}^T \frac{d|r(t)|}{d t}$ \\
    \midrule
    \makecell[l]{A \\ (average acceleration)} & mm/s2 & Rate of change of the instrument’s velocity & $ \frac{1}{T} \int_{t=0}^T \frac{d^2|r(t)|}{d t^2} $ \\
    \midrule
    \makecell[l]{MS \\ (motion smoothness)} & mm/s3 & Abrupt changes in acceleration resulting in jerky movements of the instrument & $ \sqrt{\frac{T^5}{2 \cdot P L^2} \int_{t=0}^T\left(\frac{d^3|r(t)|}{d t^3}\right)^2} $ \\
    \midrule
    \makecell[l]{EOV \\ (economy of volume)} & -     & Relationship between the maximum volume occupied by the instrument and the total path length & $ \sqrt[3]{\frac{[\underset{t}{\operatorname{Max}}(x)-\underset{t}{\operatorname{Min}(x)}] \cdot [\underset{t}{\operatorname{Max}}(y)-\underset{t}{\operatorname{Min}(y)}] \cdot [\underset{t}{\operatorname{Max}}(z)-\underset{t}{\operatorname{Min}(z)}]}{PL}} $ \\
    \bottomrule
    \end{tabular}
    }
\end{table}


\bibliographystyle{ieeetr}
\bibliography{reference}

\end{document}